\documentclass{article}

\usepackage{PRIMEarxiv}

\usepackage[utf8]{inputenc} % allow utf-8 input
\usepackage[T1]{fontenc}    % use 8-bit T1 fonts
\usepackage{hyperref}       % hyperlinks
\usepackage{url}            % simple URL typesetting
\usepackage{booktabs}       % professional-quality tables
\usepackage{amsfonts}       % blackboard math symbols
\usepackage{nicefrac}       % compact symbols for 1/2, etc.
\usepackage{microtype}      % microtypography
\usepackage{lipsum}
\usepackage{fancyhdr}       % header
\usepackage{graphicx}       % graphics

\graphicspath{{media/}}     % organize your images and other figures under media/ folder
\usepackage{authblk}
\usepackage{cite}
\usepackage{amsmath,amssymb,amsfonts}
\usepackage{algorithmic}
\usepackage{textcomp}
\usepackage{multirow}
\usepackage{hhline}
\usepackage{tabularray}
\usepackage{makecell}
\usepackage{latexsym}
\usepackage{diagbox}
\usepackage{ragged2e}

\def\BibTeX{{\rm B\kern-.05em{\sc i\kern-.025em b}\kern-.08em
    T\kern-.1667em\lower.7ex\hbox{E}\kern-.125emX}}
\begin{document}
%\title{LV-CadeNet: Clinical MEG Spike Detection with Human-Like Expertise}
\title{LV-CadeNet: A Long-View Feature Convolution-Attention Fusion Encoder-Decoder Network for EEG/MEG Spike Analysis}

% \\
% Author 1\textsuperscript{1},
% Author 2\textsuperscript{2},
% Author 3\textsuperscript{1},
% Author 4\textsuperscript{1},
% Author 5\textsuperscript{2},
% Author 6\textsuperscript{2},
% Author 7\textsuperscript{1,*}
% \\
% \bigskip
% \bf{1} Affiliation A
% \\
% \bf{2} Affiliation B
% \\

\renewcommand{\thefootnote}{}

\author[1,$\ast$]{Kuntao Xiao}
\author[2,$\ast$]{Xiongfei Wang }
\author[2]{Pengfei Teng}
\author[1]{Yi Sun}
\author[3]{Yong Zhang}
\author[1]{Wanli Yang}
\author[1]{Zikang Xu}
\author[4]{Liang Zhang}
\author[4]{Hanyang Dong}
\author[2,$\dagger$]{Guoming Luan }
\author[1,$\dagger$]{Shurong Sheng}
\affil[1]{\footnotesize the Anhui Province Key Laboratory of Biomedical Imaging and Intelligent Processing, Institute of Artificial Intelligence, Hefei Comprehensive National Science Center, Hefei, China. (email: xiaokuntao@iai.ustc.edu.cn, shengshurong@iai.ustc.edu.cn)}
\affil[2]{\footnotesize the Department of Neurosurgery, Beijing Key Laboratory of Epilepsy, Sanbo Brain Hospital Capital Medical University, Beijing, China (email: luangm@ccmu.edu.cn).}
\affil[3]{\footnotesize the Huaibei People's Hospital of Anhui Province, Huaibei, China.}
\footnotetext{$^\ast$Kuntao Xiao and Xiongfei Wang are co-first authors.}
\affil[4]{\footnotesize the AHU-IAI AI Joint Lab, School of Artificial Intelligence, Anhui University, Hefei, China.}
\footnotetext{$^\dagger$Guoming Luan and Shurong Sheng are co-corresponding authors.}

% \author{Kuntao Xiao, Xiongfei Wang, Pengfei Teng, Yi Sun, Wanli Yang, Liang Zhang, Hanyang Dong, Guoming Luan, and Shurong Sheng 
% \thanks[4]{
% This work was partly sponsored by Beijing Nova Program (20240484590), and partly supported by Beijing Natural Science Foundation (L222033). 
% Kuntao Xiao and Xiongfei Wang are co-first authors. Guoming Luan and Shurong Sheng are co-corresponding authors.}
% \thanks[5]{Kuntao Xiao, Yi Sun, Wanli Yang, and Shurong Sheng are with Anhui Province Key Laboratory of Biomedical Imaging and Intelligent Processing, Institute of Artificial Intelligence, Hefei Comprehensive National Science Center, Hefei, China. (email:  shengshurong@iai.ustc.edu.cn)}
% \thanks{Liang Zhang and Hanyang Dong are with the AHU-IAI AI Joint Lab, School of Artificial Intelligence, Anhui University, Hefei, China.}
% \thanks{Xiongfei Wang, Pengfei Teng, and Guoming Luan are with the Department of Neurosurgery, Beijing Key Laboratory of Epilepsy, Sanbo Brain Hospital Capital Medical University, Beijing, China (email: luangm@ccmu.edu.cn).}
% }

\maketitle

\begin{abstract}
The analysis of interictal epileptiform discharges (IEDs) in magnetoencephalography (MEG) or electroencephalogram (EEG) recordings represents a critical component in the diagnosis of epilepsy. However, manual analysis of these IEDs, which appear as epileptic spikes, from the large amount of MEG/EEG data is labor intensive and requires high expertise. Although automated methods have been developed to address this challenge, current approaches fail to fully emulate clinical experts' diagnostic intelligence in two key aspects: (1) their analysis on the input signals is limited to short temporal windows matching individual spike durations, missing the extended contextual patterns clinicians use to assess significance; and (2) they fail to adequately capture the dipole patterns with simultaneous positive-negative potential distributions across adjacent sensors that serve as clinicians' key diagnostic criterion for IED identification. 
To bridge this artificial-human intelligence gap, we propose a novel deep learning framework LV-CadeNet that integrates two key innovations: (1) a Long-View morphological feature representation that mimics expert clinicians' comprehensive assessment of both local spike characteristics and long-view contextual information, and (2) a hierarchical Encoder-Decoder NETwork that employs Convolution-Attention blocks for multi-scale spatiotemporal feature learning with progressive abstraction. Extensive evaluations confirm the superior performance of LV-CadeNet, which outperforms six state-of-the-art methods in EEG spike classification on TUEV, the largest public EEG spike dataset. 
Additionally, LV-CadeNet attains a significant improvement of 13.58\% in balanced accuracy over the leading baseline for MEG spike detection on a clinical MEG dataset from Sanbo Brain Hospital, Capital Medical University. 
\end{abstract}

\section{Introduction}
\label{sec:introduction}

Epilepsy stands as one of the most prevalent and serious neurological disorders worldwide \cite{Hirtz2007}, affecting approximately 50 million people by 2021 \cite{feigin2025global}. Early diagnosis is critical, as studies indicate that proper treatment can allow up to 70\% of patients to achieve seizure-free outcomes \cite{abdi2024review}. IEDs, commonly referred to as spikes, represent transient electrophysiological events that can be observed in EEG or MEG recordings and are considered key biomarkers for epilepsy \cite{barkley2003meg, ref_spikenet}. As complementary neuroimaging modalities, EEG and MEG provide distinct advantages in epilepsy evaluation: EEG measures electrical potentials through scalp electrodes, while MEG detects the associated magnetic fields using superconducting sensors, offering superior spatial resolution for deeper cortical sources. Accurate identification and analysis of IEDs play a critical role in multiple clinical aspects such as the establishment of epilepsy diagnosis, and the localization of epileptogenic zones for potential surgical intervention. However, visual analysis of MEG/EEG data with millisecond-level high time resolution for spike analysis presents significant challenges, being both labor-intensive and requiring specialized neurophysiological expertise. This substantial clinical burden underscores the importance of developing automated spike analysis systems to improve the efficiency of diagnostic workflow.  Automated spike analysis is typically formalized as a binary identification or multiclass classification problem. In binary identification, the input is classified as either a spike or non-spike, while in multi-class classification, it is categorized into specific spike types. In this study, we conduct both binary identification of MEG spikes that excels in localizing cortical sources and multiclass classification of EEG spikes that is more widely available in clinical settings.  

\begin{figure*}
\centering 
\includegraphics[width=\linewidth]{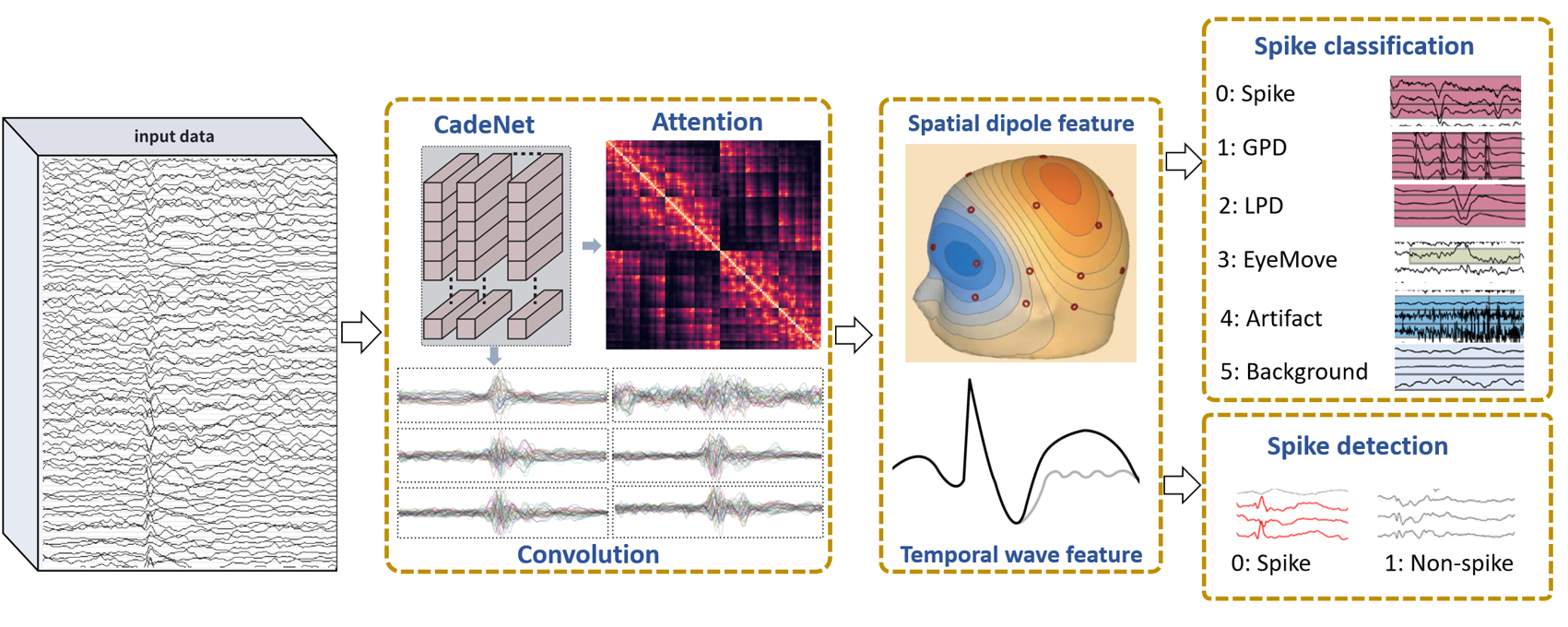}
\caption{Framework overview of CadeNet: feature extraction model we propose for spike classification and detection tasks.} 
\label{fig:task}    
\end{figure*}

Research progress in automated epileptic spike analysis from MEG/EEG data has undergone significant evolution over the past two decades. Early works in this field focus on feature engineering approaches, where manually designed features were extracted from temporal, spectral, or time-frequency domains to feed conventional machine learning classifiers including linear models and support vector machines \cite{eeg_spike_detection_svm, dumpelmann1999visual, meg_spike_detection_svm}. While these methods provided interpretable decision-making processes, their clinical applicability was substantially constrained by the requirement for meticulous parameter tuning specific to individual datasets, limited generalization capability across different datasets, and inadequate representation of complex spike morphologies. The advent of deep learning has revolutionized this field by enabling automatic feature learning directly from raw signals through end-to-end trainable architectures, achieving superior performance on large-scale datasets. These modern frameworks typically implement an integrated processing pipeline that first segments continuous signals into spike-duration clips before applying deep neural networks for feature representation learning and classification. Although pioneering deep learning studies focused primarily on temporal feature extraction \cite{EMS-Net, ref_spikenet, satelight}, growing evidence confirms that spatial dipole distribution features are as vital as temporal features \cite{MEG_Epilepsy_2021,IED_2020}, leading to the development of two distinct spatial integration paradigms: explicit multimodal fusion approaches that combine separately processed temporal signals and spatial topographies \cite{V2IED, zhang2025crossconvpyramid}, and implicit joint learning methods that employ specialized architectures to directly extract spatiotemporal features. The latter category encompasses three principal architectures: (1) convolutional neural networks that leverage their inherent spatial processing capabilities \cite{dong2024automated}, (2) attention-mechanism-based architectures that model long-range dependencies \cite{jiang2024large, wang2024eegpt}, and (3) hybrid convolution-attention architectures that synergistically combine local and global feature extraction \cite{TONG2025epilepsia}.

Despite these advancements, deep learning-based MEG/EEG spike analysis still faces two major challenges. First, the prevalent segment-based analysis paradigm restricts models' access to broader electrophysiological context, causing frequent misclassification of ambiguous waveforms that require long-recording evaluation. Second, current spatiotemporal feature extraction methods present inherent limitations: explicit multimodal fusion approaches suffer from high computational costs and feature alignment issues, while more efficient implicit joint learning methods lack multi-scale analysis capabilities and dedicated feature decoding modules. These technical shortcomings lead to compromised clinical performance, including higher false positive rates, and limited generalizability across different recording configurations.

To jointly address these two challenges, we propose LV-CadeNet, a novel deep learning framework that incorporates two key innovations. First, we have developed six input features that quantify the morphological significance of signal segments across extended time windows. By combining these enhanced significance features from a broader temporal perspective with raw input segments, our approach effectively captures spike characteristics across wider temporal spectra while minimizing the incorporation of extraneous noise. This methodology emulates the meticulous annotation process of clinical experts, who typically identify MEG spikes ranging from 27 to 120 ms \cite{Nowak_Santiuste_Russi_2009} within a 5-10 second signal context. We term this strategy as ``long-view input feature construction" throughout this study. Second, we introduce a novel \textbf{C}onvolution-\textbf{a}ttention fusion \textbf{e}ncoder-\textbf{d}ecoder \textbf{N}etwork (CadeNet) for EEG/MEG spike analysis. The encoder employs multiple processing blocks to capture multi-scale representations with progressive abstraction. Within each block, temporal convolution operations are employed for precise temporal waveform characterization, and attention mechanisms are utilized for context-aware spatial feature extraction across sensor arrays. The decoder implements an adaptive feature distillation process that dynamically weights the representation obtained from the encoder, effectively emulating the selective attention mechanisms employed by clinical experts during visual analysis. This biologically-inspired design enables the network to automatically focus on the most discriminative spatiotemporal patterns while suppressing irrelevant background activity, thereby improving spike analysis performance. To enhance intuition, Fig. \ref{fig:task} depicts the core idea behind CadeNet and the tasks addressed in this work. The principal contributions of our work include:

\begin{itemize}
    \item We construct long-view input features that comprehensively capture the morphological significance of signal segments while preserving the original raw input through two key innovations:(1) a matrix-based representation of morphological features that enables seamless integration with raw signal input and (2) a novel context normalization technique that enhances feature discriminability across extended temporal scales. To our knowledge, this is the first systematic approach to quantify and incorporate the long-range morphological context in EEG/MEG spike analysis.
    This method demonstrates consistent performance gains across diverse network architectures, confirming its robustness and generalizability.
    \item We present a novel encoder-decoder architecture designed for comprehensive spatiotemporal feature extraction. The encoder captures multi-scale representations with progressive abstraction while the decoder performs adaptive feature decoding for accurate spike analysis. By progressively increasing the abstraction of extracted spatiotemporal features and performing feature decoding, our architecture overcomes fundamental limitations of conventional approaches and achieves state-of-the-art performance in both EEG and MEG spike analysis tasks.
    \item Extensive experiments have been conducted to benchmark our approach against current state-of-the-art spike analysis techniques, and ablation studies have been developed to prove the robustness of the framework. 
\end{itemize}

The remainder of this paper is organized as follows. Section II reviews related research. Next, Section III describes our methodology. Section IV discusses the experiments and the results of the evaluation. Finally, Section V concludes this paper and provides directions for future research.

\section{Related Work}
The initial development of automatic spike analysis was accomplished by Stevens et al. in 1972 \cite{automatic_spike_detection_stevens}, who employed power density spectrum analysis to isolate epileptiform spikes from prolonged EEG recordings. Following this, pioneering contributions such as \cite{spike_detection_poineerwork} proposed by Gotman et al., identified epileptic spikes with a particular emphasis on the differentiation of non-epileptic transients tailored to the subject’s state, including active or quiet wakefulness and various sleep stages. Subsequently, traditional machine learning methods such as support vector machines \cite{eeg_spike_detection_svm, meg_spike_detection_svm}, and linear modeling \cite{dumpelmann1999visual} have been utilized to different forms of the input EEG signals for automatic epileptic spike detection. Meanwhile, shallow neural network based spike detection method such as \cite{ref_spike_detection_2016} emerged as an important branch in this field.

In recent years, deep learning has emerged as the dominant paradigm for automatic epileptic spike analysis, leveraging its powerful capacity to learn complex patterns from neurophysiological data. Early approaches primarily focused on temporal feature extraction \cite{EMS-Net, ref_spikenet, satelight}. Among these studies, the SpikeNet model proposed in \cite{ref_spikenet} is adapted from ResNet architectures \cite{ref_resnet} for spike detection, and Satelight \cite{satelight} combines convolutional layers with self-attention mechanisms for temporal feature extraction. As research advanced, the importance of spatial features became evident, leading to specialized spike detection models that jointly model temporal and spatial characteristics \cite{V2IED, zhang2025crossconvpyramid, dong2024automated}. The CrossConvPyramid approach \cite{zhang2025crossconvpyramid} processes both topographic images and MEG signals for MEG spike detection, and \cite{TONG2025epilepsia} employs hybrid architectures integrating spatial attention with temporal convolutions for EEG spike detection. More recently, the field has witnessed the rise of general-purpose foundation models that handle spike analysis alongside other neurophysiological tasks through transformer-based architectures and unsupervised pretraining \cite{BIOT, eeg2rep2024, jiang2024large
, wang2024eegpt}. These foundation models, exemplified by BIOT \cite{BIOT}, employ a two-stage learning paradigm: first learning general representations of EEG through unsupervised pre-training that transforms heterogeneous biosignals with varying channel configurations, temporal lengths into standardized sentence-like structures, followed by task-specific fine-tuning for spike classification. This approach enables the model to leverage broad electrophysiological patterns learned from diverse data before specializing for the target application.

\section{Materials and method}
\subsection{Datasets and pre-processing approaches}
\subsubsection{Datasets}
This study evaluates LV-CadeNet on two clinical tasks: EEG and MEG spike analysis. For EEG analysis, we utilize the publicly available Temple University Hospital EEG Events Corpus (TUEV) \cite{THU2016}, which comprises scalp EEG recordings from 370 patients collected using the standard 10-20 international electrode placement system. Expert neurophysiologists at Temple University Hospital meticulously annotated six distinct waveform patterns within 1-second bipolar channel segments: (1) spike-and-wave complexes (SPSWs), (2) generalized periodic discharges (GPDs), (3) lateralized periodic discharges (LPDs), (4) eye movement artifacts, (5) technical artifacts, and (6) background activity (BCKG). The first three categories represent pathological epileptiform activity, while the remaining three serve as negative controls. Since there are six categories within this dataset, our task on this dataset is formalized as a multiclass classification problem. For data separation, we adhered to the predefined training-test split specified in the original dataset, with statistics of segmented samples for each category detailed in Table \ref{tab:dataset1}. 

\begin{table}[t]
 \vspace{-1 em}
 \caption{Statistics of the datasets for EEG spike classification}
    \centering
    \resizebox{5cm}{1.5cm}{
    \begin{tabular}{c|c|c}
    \hline
         \diagbox[]{Event}{Dataset} & Train & Test \\ \hline
         SPSWs & 645 & 329  \\ \hline
         GPDs &  11254 & 4677   \\ \hline
         LPDs & 6184 &  1998  \\ \hline
         Eye movement & 1070 & 567   \\ \hline
         Technical artifacts & 11053 &  2204  \\ \hline
         BCKG &  53726 & 19646   \\ 
    \hline
    \end{tabular}} 
    \label{tab:dataset1}
     \vspace{-1em}
\end{table}

\begin{table}[t]
 \caption{Statistics of the datasets for MEG spike detection}
    \centering
    \scalebox{0.8}{
        \begin{tabular}{c|c|c|c|c}
        \hline
             \diagbox[]{Dataset}{Info} & Subjects & files & Spikes & Non-Spikes \\ \hline
             Train & 119 & 468 & 14780 & 17511 
             \\ \hline
             Test & 50 & 50 & 3513 & 115001   \\ 
        \hline
        \end{tabular}
    }
    \label{tab:dataset2}
    \vspace{-2em}
\end{table}

For MEG spike detection, we employ two proprietary datasets: Sanbo-CMR for training and Sanbo-Clinic for evaluation. Sanbo-CMR, jointly established by Sanbo Brain Hospital, Capital Medical University and Peking University's Center for MRI Research, contains MEG recordings with expert-annotated spikes and non-spikes in a near-balanced distribution, consistent with previous MEG spike detection studies. Sanbo-Clinic, collected exclusively at Sanbo Brain Hospital, features comprehensive neurophysiologist annotations that reflect clinical reality, with a natural imbalance between positive and negative samples. Complete dataset statistics are provided in Table \ref{tab:dataset2}. The study protocol was approved by both institutional review boards\footnote{(1) Sanbo MEG dataset: Institutional Review Board (IRB) of Capital Medical University (Protocol numbers: SBNK-YJ-2023-002-03); (2) CMR MEG dataset: IRB of Peking University, Beijing, China (Protocol number: \#2017-08-01). }, with written informed consent obtained from all participants.

In subsequent methodological presentation, we primarily utilize epileptic EEG spike classification as the foundational paradigm to elucidate core concepts and technical implementations. The MEG spike detection task is strategically reserved as an independent validation modality, serving to rigorously demonstrate the robustness and generalizability of our framework in the results section.

\subsubsection{Preprocessing approaches} We adopted standardized preprocessing pipelines for both EEG and MEG data in accordance with previous studies. The preprocessing sequence consisted of three key steps: (1) Bandpass filtering was applied to extract physiologically relevant frequency components (EEG: 0.1-75.0 Hz; MEG: 3-40 Hz), followed by a 50 Hz notch filter to eliminate power-line interference; (2) all recordings were downsampled to 200 Hz (EEG) and 250 Hz (MEG) to enhance computational efficiency while preserving diagnostically relevant information;
(3) we performed modality-specific normalization to ensure consistent feature scaling across all recordings 
by computing file-wise z-scores using each recording's mean amplitude and standard deviation.

It is worth noting that prior to implementing our core processing pipeline, we performed modality-specific signal transformations to optimize data quality. For EEG recordings, we enhanced spatial information by systematically computing bipolar derivations between multiple electrode pairs, transforming the original signal dimensions from ($\hat{C}$, $T$) to an enriched representation ($C$, $T$) where $C > \hat{C}$. This potential difference calculation approach, implemented through multiple montage configurations, would significantly improve noise resilience. In parallel, MEG data underwent specialized artifact reduction processing, beginning with Maxwell filtering \cite{taulu2006spatiotemporal} to eliminate environmental magnetic interference, followed by independent component analysis (ICA) to identify and remove physiological artifacts including ocular movements (EOG) and cardiac signals (ECG). 

\subsection{LV-CadeNet: Long View feature Convolution-Attention fusion Encoder-Decoder Network}

\begin{figure*}
\centering 
\includegraphics[width=\linewidth]{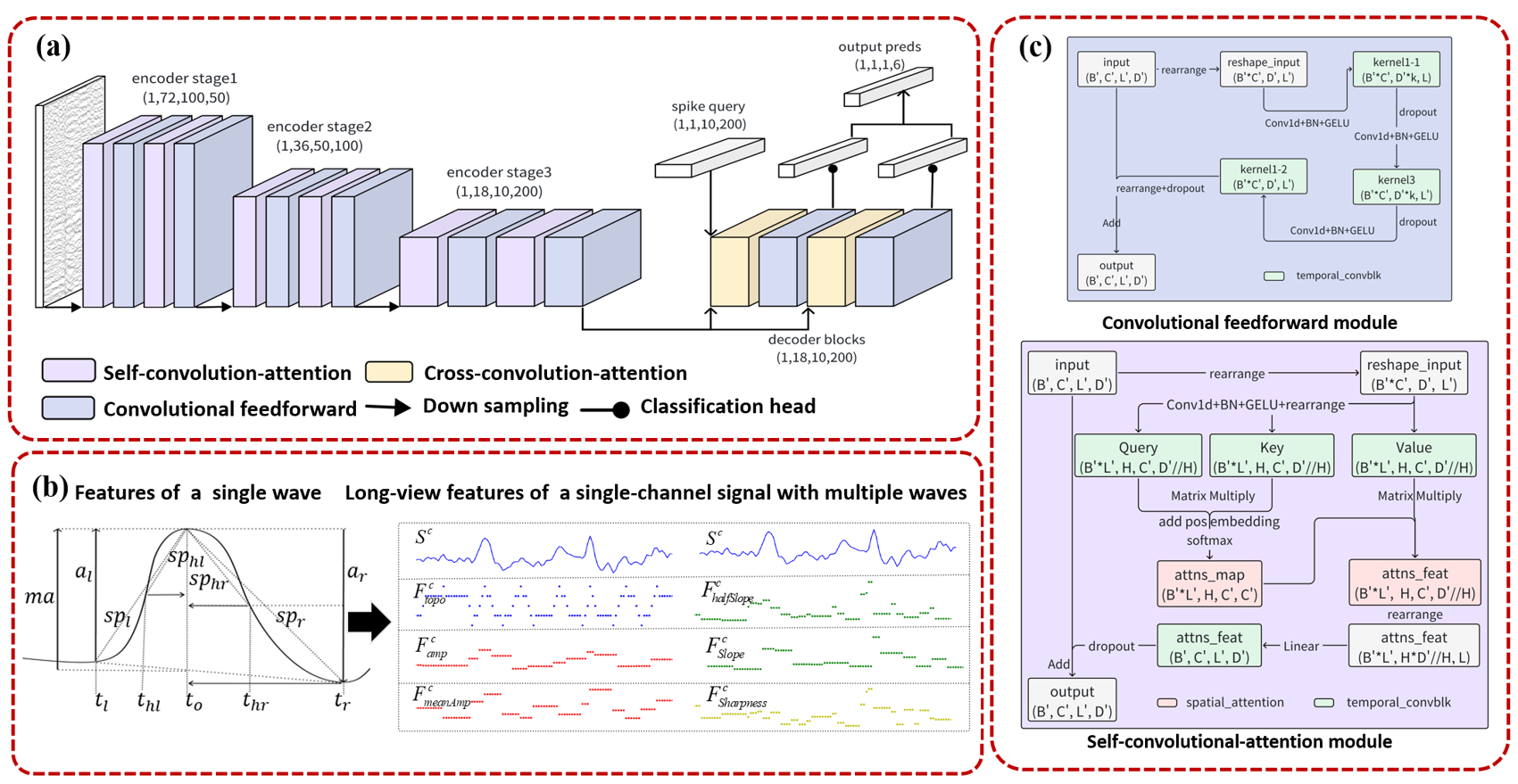}
\caption{(a) Architecture of our feature extraction model CadeNet when the batch size is 1. (b) Illustration of the long-view input features fed into CadeNet. (c) Structure of key modules in CadeNet: cross-convolution-attention module omitted due to similarity with self-convolution-attention.} 
\label{fig:framework}    
\end{figure*}

For a preprocessed EEG signal file $\mathcal{S} \in \mathbb{R}^{C\times T}$, representing the $C$-channel signal variations across a time duration of $T$, the objective of our LV-CadeNet approach is to classify each signal segment $S$ within $\mathcal{S}$ to its belonging spike types. To achieve this, we first construct the long view input features $\mathcal{F}$ for $\mathcal{S}$. Subsequently, we propose an advanced convolution-attention fusion model, CadeNet, to extract features from $\mathcal{X}=[\mathcal{S},\mathcal{F}]$, and harbors spike types using a classification head. An illustration of LV-CadeNet is presented in Fig. \ref{fig:framework}, with a detailed explanation of its components provided in the subsequent two sections.

\subsubsection{Long view input feature construction}
 To generate the long view input feature $F$ for a given input signal segment $S$, we initiate the process by leveraging the file-level signal $\mathcal{S}$ to derive the file-level feature $\mathcal{F}$ with the four-step process outlined below.
 
 Firstly, we identify the local maxima and minima in $\mathcal{S}$ through extremum detection, considering a complete wave with a maximum moment $t_o$ between two minima $t_{l},t_{r}$ as a complete wave \cite{Cui_Cao_Wang_Zheng_Cen_Teng_Luan_Gao_2022}. Using this method, the signal of each channel $\mathcal{S}^c$ where $c$ ranges from 0 to $C$ can be divided into a certain number of complete waves.

 \begin{align}
    a_{l} &= \mathcal{S}^{c,t_o} - \mathcal{S}^{c,t_{l}}; 
    a_{r} = \mathcal{S}^{c,t_{r}} - \mathcal{S}^{c,t_o} \\
    {sp}_{l} &= \frac{a_l}{t_o - t_{l}};  {sp}_{r} = \frac{a_r}{t_{r} - t_o}\\
    hsp_{l} &= \frac{a_l}{2\cdot(t_o - t_{hl})}; hsp_{r} = \frac{a_r}{2\cdot(t_{hr} - t_o)} \\
    ma &= \frac{a_{l} \cdot (t_{r} - t_o) + a_r \cdot (t_o - t_{l})}{t_{r}-t_{l}} \\
    sn &= \left | \textstyle \sum^{t_{o}+4}_{t=t_{o}-4} \mathcal{S}^{c,t} - 9 \cdot \mathcal{S}^{c,t_o} \right |
    \label{eq_finalfea}
\end{align}
 Secondly, we construct six features for each signal $\mathcal{S}^{c}$. The first feature depicting topological structure of $\mathcal{S}^c$ is denoted as $\mathcal{F}^c_{topo}$. More specifically, signals at local maxima are assigned a value of 1 for $\mathcal{S}^c$. Conversely, signals at local minima are assigned a value of -1. For signals occurring at the half-width high points, which lie between local maxima and minima, a value of 0 is attributed. Signals at other time points are assigned values of 0.5 or -0.5, depending on their proximity to the local maxima or minima, respectively. The other five features are represented as $\mathcal{F}^c_{amp}$, $\mathcal{F}^c_{meanAmp}$, $\mathcal{F}^c_{slope}$, $\mathcal{F}^c_{halfSlope}$ and $\mathcal{F}^c_{sharpness}$, each providing insights into a distinct morphological property of $\mathcal{S}^{c}$. To construct these features, we first calculate key properties for a single complete wave. Taking the wave with a time ranging from $t_{l}$ to $t_o$ to $t_{r}$ within $\mathcal{S}^c$ as an example, we define the left and right adjacent half-width high moments as $t_{hl}$ and $t_{hr}$ respectively. Then the basic properties of this wave can be computed with Equations (1)-(5). In these formulas, $a$, $sp$, $hsp$, and $sn$ represent the abbreviations for the amplitude, slope, half-width slope and sharpness of the wave respectively \cite{Nowak_Santiuste_Russi_2009}, while $ma$ refers to the mean amplitude of the wave. ``$l$" is used to denote properties of the left half of the wave, and ``$r$" is used to denote properties of the right half of the wave. Utilizing these essential wave characteristics, we can construct the five features derived from Equations (1)-(5) that align with the shape of the original $\mathcal{S}^c \in \mathbb{R}^{1 \times T}$ by interpolating the available features across the time intervals. Specifically, for each single wave, the elements spanning from $t_{hl}$ to $t_{o}$ within matrix $\mathcal{F}^{c}_{amp}$ are denoted as $\mathcal{F}^{c,t_{hl}:t_{o}}_{amp}$ and assigned the value $a_l$. In alike manner,  $\mathcal{F}^{c,t_{o}:t_{hr}}_{amp}$ receives the value $a_r$. For the feature related to mean amplitude, $\mathcal{F}^{c,t_{hl}:t_{hr}}_{meanAmp}$ is given the value $ma$. Regarding the slope-related feature, $\mathcal{F}^{c,t_{hl}:t_{o}}_{slope}$ acquires the value $sp_l$, and $\mathcal{F}^{c,t_{o}:t_{hr}}_{slope}$ acquires the value $sp_r$. For the feature related to half slope, $hsp_{l}$ is attributed to $\mathcal{F}^{c,t_{hl}:t_{o}}_{halfSlope}$ and $hsp_{r}$ is attributed to $\mathcal{F}^{c,t_{o}:t_{hr}}_{halfSlope}$. $\mathcal{F}^{c,t_{hl}:t_{hr}}_{sharpness}$ is assigned the value $sn$.

Thirdly, we perform long-view normalization on the latter five of the six aforementioned features. For each feature $\mathcal{F}^{c}_{*}$ among these five, we compute the z-score of its subset $\mathcal{F}^{c, t_r-t_l}_{*}$, which represents a complete wave, using the distribution of values from the surrounding 100 complete waves. This neighboring window spans approximately 6 seconds, substantially longer than the duration of a single wave ($t_r - t_l$). 
This 6-second interval is selected according to the signal perspective essential for the manual annotation of spikes by our participating clinical experts. Since the feature construction approaches are applied to each $\mathcal{S}^{c}$, the resulting features are file-level features denoted as $\mathcal{F}_{topo}$, $\mathcal{F}_{amp}$, $\mathcal{F}_{meanAmp}$, $\mathcal{F}_{slope}$, $\mathcal{F}_{halfSlope}$, $\mathcal{F}_{sharpness}$ respectively, each having a dimension of (C, T).
 
Finally, we concatenate the six long view features introduced above with the signal $\mathcal{S}$ to form a long view feature matrix $\mathcal{X} \in \mathbb{R}^{C\times T\times 7}$ at the file-level. The input slice $X^i \in \mathbb{R}^{C\times L\times 7}$ is extracted by selecting the signal centered on the annotation point $i$ with a duration of $L$ from the file matrix $\mathcal{X}$. An intuitive diagram depicting the input $X^i$ is presented in Fig. \ref{fig:framework}. Subsequently, we aggregate a collection of slices into a batch represented as $X = [X^b| b \in[1, B]]\in \mathbb{R}^{B\times C\times L\times D}$ for training where $B$ represents the batch size. In the subsequent section of this research, we refer to the first dimension of a feature matrix as the batch dimension, the second as the spatial or sensor dimension, the third as the temporal dimension, and the fourth as the feature dimension. This nomenclature is adopted to clarify and enhance the explanation of our methodology. 

\subsubsection{The Convolution-attention fusion encoder-Decoder Network}

The proposed CadeNet adopts an encoder-decoder architecture specifically designed for efficient spatiotemporal feature learning and classification. The three-stage encoder extracts features from the input $X$ with progressively increasing abstraction, while the single-stage decoder strategically compresses the features into a compact latent space optimized for classification. 
The subsequent discussion will explore the integral components that make up CadeNet. Within these components, a residual connection is integrated into each submodule within the convolution-attention and convolutional feedforward blocks. As a result, the output maintains the same dimensions as the input for these two blocks, ensuring a seamless flow of information.

\noindent \textbf{Convolution-Attention Block.} The convolution-attention block in CadeNet includes two parts: the self-convolution-attention module and the cross-convolution-attention module. For the self-convolution-attention module, it's input $Z_{slf}$ is a feature either transformed from the input $X$ or derived from the Downsample module of CadeNet. Suppose $Z_{slf}$ has a dimension of $(\hat{B}, \hat{C}, \hat{L}, \hat{D})$, the following equations will be applied to $Z_{slf}$ to extract temporal and spatial features:
\begin{align}
    Z_1 &= ConvBlk(Z_{slf} \xrightarrow{Trans} \mathbb{R}^{(\hat{B} \cdot \hat{C}) \times \hat{D} \times \hat{L}}), \\ 
    \notag Z_1  &\in \mathbb{R}^{(\hat{B} \cdot \hat{C}) \times (3 \cdot H_{slf} \cdot D_{slf}) \times \hat{L}}\\
    Z_2 &= Z_1 \xrightarrow{Trans} \mathbb{R}^{3 \times (\hat{B} \cdot \hat{L}) \times H_{slf} \times \hat{C} \times D_{slf}} \\
    Q_{slf} &= Z_2[0, :]; K_{slf} = Z_2[1, :]; V_{slf} = Z_2[2, :] \\
    Self&Attention(Z_2) = SoftMax(\frac{Q_{slf} \ast K_{slf}}{\sqrt{D_{slf}}}) \ast V_{slf}\\
    Z_3 &= MultiHead(SelfAttention(Z_2)),\\
    \notag Z_3 &\in \mathbb{R}^{(\hat{B} \cdot 
    \hat{L}) \times H_{slf} \times \hat{C} \times D_{slf}} \\
    Z_4 &= ConvBlk(Z_3 \xrightarrow{Trans} \mathbb{R}^{(\hat{B} \cdot \hat{C}) \times (H_{slf} \cdot D_{slf}) \times \hat{L}})\\
    Z'_{slf} &= Z_{slf} + (dropout(Z_4) \xrightarrow{Trans} \mathbb{R}^{\hat{B} \times \hat{C} \times \hat{L}  \times \hat{D}})
\end{align}

Where the $ConvBlk$ in the equations consists of a 1D convolutional layer operated along the temporal dimension, followed by a batch normalization layer \cite{Ioffe_Szegedy_2015} and GELU activation layer\cite{Hendrycks_Gimpel_2016}. The $SelfAttention$ and $MultiHead$ are referenced from \cite{transformer2017}. $H_{slf}$ and $D_{slf}$ represent the dimensionality for computing the query matrix $Q_{slf}$, the key matrix $K_{slf}$, and the value matrix $V_{slf}$. $dropout$ \cite{Srivastava_Hinton_Krizhevsky_Sutskever_Salakhutdinov_2014} is a function that prevents the convolution-attention block from overfitting. In subsequent experiments, the value of $H_{slf}$ is fixed at 5. $D_{slf}$ is determined through the division of $\hat{D}$ and $H_{slf}$. 

For the cross-convolution-attention module, unlike the self-convolution-attention module which processes a single input, this module takes two inputs: the feature matrix $Z_{cros}$, which is the output from the encoder at a particular stage, and a learnable external feature matrix $Z_{query}$. The computation of the output for this module follows a similar approach to the self-convolution-attention module, with one key distinction: the Query vector $Q_{cros}$, which serves as an input to the multihead attention layer, is derived from the external feature matrix $Z_{query}$, analogous to how $Q_{slf}$ is derived from $Z_{slf}$. Concurrently, the Key and Value vectors $K_{cros}$ and $V_{cros}$, which are the other two inputs to the multihead attention layer, are obtained in the same manner as the self-convolution-attention module, from the encoder-produced feature matrix $Z_{cros}$.

\noindent \textbf{Convolutional Feedforward Block.} The convolutional feedforward module is utilized at every stage within both the encoder and decoder, subsequent to the convolution-attention module. This module intricately processes the feature derived from a particular convolution-attention block along the temporal axis. The input feature for this block, denoted as $Z_{ff} \in \mathbb{R}^{\hat{B} \times \hat{C} \times \hat{L} \times \hat{D}} $ retains the same dimensions as the input to the convolution-attention block due to the residual connections between submodules. To refine the temporal characteristics, we transform $Z_{ff}$ to a shape of $(\hat{B} \cdot \hat{C}, \hat{D}, \hat{L})$. Subsequently, three $ConvBlk$s with $dropout$ are applied to this transformed version of $Z_{ff}$. The process concludes with a residual connection linking the input and output. The formal mathmatical presentation of this block is provided in Equations (13)-(15).
\begin{align}
    Z_5 &= Z_{ff} \xrightarrow{Trans} \mathbb{R}^{(\hat{B} \cdot \hat{C}) \times \hat{D} \times \hat{L}}\\
    Z_{conv} &= (\prod^{3}  dropout\odot ConvBlk) (Z_5) \\ 
    Z'_{ff} &= Z_{ff} + Z_{conv}\xrightarrow{Trans} \mathbb{R}^{\hat{B} \times \hat{C} \times \hat{L} \times \hat{D}}
\end{align}

\noindent\textbf{Downsample Block.} In this block, the downsampling layer is designed to condense the features $Z_{ds} \in \mathbb{R}^{\hat{B} \times \hat{C} \times \hat{L} \times \hat{D}}$, which is the output of the convolutional feedforward block. It targets both the spatial and temporal dimensions within the encoder, shifting information from these two dimensions to the feature dimension, which is essential for the classification in the decoder. The layer converts $Z_{ds}$ to $Z'_{ds} \in \mathbb{R}^{\hat{B} \times \frac{\hat{C}}{p} \times \frac{\hat{L}}{q} \times (\hat{D} \cdot p \cdot q)}$. This information swap between different dimensionalities is achieved with two steps: first, the Pixel Shuffle method \cite{pixelshuffle2016} is applied to compress the spatial or temporal data and integrate it into the feature dimension, followed by a one-dimension convolutional layer that adjusts the last feature dimension to desired size. It is worth noting that the parameter $p$ corresponds to the number of sensor groups generating the EEG/MEG signals, with sensors in close proximity within each group. This approach results in a more coherent and interpretable representation of the spatial information within the feature dimension.

\noindent \textbf{Classification Head.} 

We utilize features $\{Z^j|j=[1,M]\}$ extracted from the convolutional feedforward module of the decoder for classification, where $j$ indexes a specific convolutional feedforward module, and $M$ denotes the number of blocks within the decoder. For each feature map $Z^j$, we apply max pooling and mean pooling along its temporal and spatial dimensions. The pooled features are subsequently transformed through a fully connected layer followed by softmax activation to generate prediction $P^j$.

Given the true label vector $Y^b$ corresponding to input $X^b$, the final prediction $P^{b,n}$ is computed as a weighted summation of all $P^j$ predictions. The learning objective of CadeNet is then formulated as the cross-entropy loss between the predicted probabilities and ground truth labels. This process can be formally expressed as follows:

\begin{align}
    P^{b,n} &= SoftMax(\sum^{M}_{j=1} w_j\cdot FC(GP(Z^j))) \\
    l_{ce} & = \frac{-1}{B}\sum^{\hat{B}}_{b=1}\sum^{\hat{N}}_{n=1}((1-\varepsilon)Y^{b,n} + \frac{\varepsilon}{2}) log(P^{b,n})
\end{align}
Here, $FC$ represents fully connected layer, $GP$ denotes the concatenation of mean pooling and max pooling operations across both spatial and temporal dimensions in Eq. (16). For Eq. (17), $\varepsilon$ corresponds to a label smoothing coefficient of 0.1 \cite{Szegedy_Vanhoucke_Ioffe_Shlens_Wojna_2016}, while $\hat{N}$ specifies the total number of categorical classifications in the task.

\section{Experiments and Results}
\subsection{Baseline models and experimental settings}
\subsubsection{Baseline models}
This study presents a systematic performance comparison between our LV-CadeNet framework and existing state-of-the-art approaches for EEG and MEG spike analysis. Our comparative analysis focuses on models with similar task objectives and reproducible implementation characteristics to ensure meaningful benchmarking results.

To ensure thorough evaluation across modalities, we compare LV-CadeNet against three categories of baseline models: (1) two EEG-specific spike classification models SpikeNet \cite{ref_spikenet} and ET-Mamba \cite{et-mamba}, (2) two MEG-specific spike detectors EMSNet \cite{EMS-Net} and FAMED \cite{FAMED}, and (3) four shared models that can be applied to both EEG and MEG. Among the four shared models, Labram \cite{jiang2024large} and EEGPT \cite{wang2024eegpt} represent foundation models designed for both modalities though initially validated only on EEG data. The remaining two models, KAN2EEG \cite{herbozo2025kan} and SimBA \cite{patro2024simba}, leverage the popular and effective KAN (Kolmogorov-Arnold Networks) \cite{liu2025kan} and Mamba \cite{gu2024mamba} architectures respectively. These two recently developed network structures have demonstrated remarkable success across various domains due to their superior modeling capacity, making them particularly suitable as shared models for our EEG and MEG spike analysis tasks. This multi-faceted comparison approach provides meaningful performance assessments against both modality-specific and general spike analysis solutions. 

\noindent \textbf{(1) EEG-specific baseline approach}
 \begin{itemize}
     \item SpikeNet \cite{ref_spikenet}: an EEG spike analysis model influenced by the architecture of residual networks and employed CNNs to adeptly extract both temporal and spatial features.
     \item ET-Mamba \cite{et-mamba}: an EEG spike classification model that first transforms EEG signals into the frequency domain and constructs an energy-based topology using channel-wise spectral power. Features are then propagated bidirectionally along energy gradients, enabling the Mamba model to capture temporal and spatial dependencies.
 \end{itemize}
\noindent \textbf{(2) MEG-specific baseline approaches}
\begin{itemize}
    \item EMS-Net \cite{EMS-Net}: a pioneering model leveraging CNNs for automatic MEG spike detection. It was characterized by the fusion of local and global temporal features, extracted by the CNNs;
    \item FAMED \cite{FAMED}: A model designed for both MEG spike classification and segmentation. It draws inspiration from the U-Net \cite{li2018deepunet} design, where both the encoder and the decoder are constructed from CNNs fortified with squeeze-and-excitation blocks that can recalibrate CNN filter responses. We follow the original work to use the encoder for MEG spike detection task in this study.
\end{itemize} 
\noindent \textbf{(3) Shared baseline approaches} 
\begin{itemize}
\item Labram \cite{jiang2024large}: a foundation model that performs cross-dataset learning for EEG signals by first segmenting EEG signals into channel patches. Subsequently, it develops a neural tokenizer using vector-quantized spectrum prediction to encode raw EEG patches into compact neural codes. Finally, it pre-trains transformer models via masked patch prediction of these neural codes. This method creates semantically meaningful representations of input data while enabling effective transfer learning across different datasets.
\item EEGPT \cite{wang2024eegpt}: a foundation model pretrained on diverse EEG datasets to maximize its adaptability across different BCI tasks. It performs a mask-based dual self-supervised learning with spatio-temporal representation alignment, which operates on EEG features with high signal-to-noise ratio instead of raw signals to ensure robust feature extraction. In addition, it uses a hierarchical architecture that separately processes spatial and temporal information for enhanced computational efficiency. 
\item KAN2EEG \cite{herbozo2025kan}: A shallow KAN-based neural network originally used for seizure detection. It works by replacing the fixed activation function in the hidden layers of the multilayer perceptrons (MLPs) with dynamic learnable activation functions founded on the Kolmogorov-Arnold representation theorem, which posits that any multivariate continuous function can be broken down into a finite compostion of continuous univariate functions and addition operation \cite{liu2025kan}. 
\item SimBA \cite{patro2024simba}:  A Mamba-based neural network originally used for multivariate time-series predictions. Instead of using attention networks for sequence mixing and MLPs for channel mixing, this novel architecture incorporates Mamba for sequence modeling and introduces EinFFT as a new channel modeling technique. It therefore addressed the inductive bias and computational complexity of transformer \cite{transformer2017} based models. 
\end{itemize}

It is worth noting that to ensure a comprehensive and fair comparison between our LV-CadeNet and existing foundation models, we have established rigorous evaluation protocols tailored to each task. For EEG spike classification, where there are multiple variants of the pre-trained foundation models, we perform extensive comparisons across different configurations. For example, we evaluate our model against Labram in various forms: the small-sized base variant with both random initialization and pretrained weights, as well as its large pretrained version. Furthermore, to provide a thorough performance analysis, we assess both compact and full-scale versions of our own architecture. In the case of MEG spike detection, where pretrained foundation models are unavailable, we ensure fair comparisons by matching model parameter scales to conventional foundation model sizes while excluding pretrained weights from all evaluations. This systematic approach allows for meaningful benchmarking across different model architectures and training paradigms.

\subsubsection{Experimental settings} 
We evaluated our LV-CadeNet and all baseline models through rigorous cross-validation for both EEG spike classification and MEG spike detection using respective datasets introduced in Section II. Specifically, we repeated each experiment 30 times with independently sampled validation sets drawn from the corresponding training data, while maintaining consistent predetermined validation splits across all trials to ensure comparability. The validation sets served only for checkpoint-based model selection without contributing to parameter optimization. The performance of the model was ultimately evaluated on the respective test sets using the optimal checkpoints selected during the validation. We implemented an early stopping criterion that terminated training if validation performance failed to improve for five consecutive epochs. All models were optimized using AdamW with an initial learning rate of 5e-4, which was decayed by a factor of 0.5 every five epochs.

\noindent\textbf{Evaluation approaches.} This study evaluates model performance using four key metrics: balanced accuracy (BAcc) to account for class imbalance, Cohen's Kappa (CKap) to measure classification agreement beyond chance, weighted F1 score (WF1) to balance precision and recall across classes, and overall accuracy (Acc) as a baseline measure of prediction correctness. These metric abbreviations BAcc, CKap, WF1, Acc are consistently used throughout the results section for clarity and conciseness. For EEG spike type classification and MEG spike detection tasks, we select a corresponding subset from the aforementioned four metrics for evaluation, ensuring consistency with the evaluation criteria used in baseline methods.

\subsection{Results and Analysis}

\begin{table}[t]
\caption{Evaluation results of different models for the EEG spike classification task.}
\label{tb:results-eeg}
\centering
\scalebox{0.75}{
    \begin{tabular}{c|ccccc}
     \hline
     \diagbox[]{Model}{Metric} & BAcc (\%) & CKap (\%) & WF1 (\%) & Prams (M)\\ \hline

      SpikeNet \cite{ref_spikenet} &  46.34 $\pm$ 2.13 & 47.12 $\pm$ 3.36 & 74.48 $\pm$ 1.40 & 6.51 \\
     ET-Mamba \cite{et-mamba} & 45.28 $\pm$ 1.39   & 43.97 $\pm$ 0.63 & 71.80 $\pm$ 0.52 & 0.83\\
     EEGPT-Tiny (rand) \cite{wang2024eegpt} & 49.64 $\pm$ 2.24  & 54.07 $\pm$ 3.19 & 77.54 $\pm$ 1.41 & 4.7  \\
     EEGPT (pretrained) \cite{wang2024eegpt}  & 62.32 $\pm$ 1.14  & 63.51 $\pm$ 1.34 & 81.87 $\pm$ 0.63 & 25  \\
     Labram-base (rand) \cite{jiang2024large} & 55.47 $\pm$ 3.35 &  56.28 $\pm$ 4.06 &  78.13 $\pm$ 1.99 & 5.8 \\
     Labram-base (pretrained)  \cite{jiang2024large}  & 64.09 $\pm$ 0.65 &  66.37 $\pm$ 0.93 &  83.12 $\pm$ 0.52 & 5.80 \\
     Labram-large (pretrained) \cite{jiang2024large}  & 66.16 $\pm$ 1.70 &  67.45 $\pm$ 1.95 &   83.29 $\pm$ 0.86 & 369 \\
     SimBA \cite{patro2024simba}  & 46.96 $\pm$ 1.92 & 49.99 $\pm$ 2.33 & 75.48 $\pm$ 1.02 & 25.8 \\
      KAN2EEG \cite{herbozo2025kan}  & 41.99 $\pm$ 1.04 & 41.10 $\pm$ 1.35 & 70.81 $\pm$ 0.69& 6.5  \\

     \hline
     CadeNet (ours)  & 64.55 $\pm$ 2.41 & 63.96 $\pm$ 2.44 & 81.74 $\pm$ 1.32 & 5.87 \\
    LV-CadeNet (ours)  & 66.83 $\pm$ 2.47 & 67.40 $\pm$ 3.72 & 83.35 $\pm$ 2.07 & 5.87  \\
    LV-CadeNet-large (ours)  & \textbf{67.32} $\pm$ 1.58 & \textbf{68.23} $\pm$ 3.41 & \textbf{83.61} $\pm$ 1.99 & 11.79 \\ \hline
    \end{tabular} 
 }
\end{table}

\begin{table}[t]
\caption{Evaluation results of different models for the MEG spike detection task.}
\label{tb:results-meg}
\centering
\scalebox{0.7}{
    \begin{tabular}{c|cccccc}
     \hline
     \diagbox[]{Model}{Metric} & BAcc (\%) & CKap (\%) & WF1 (\%) & Acc (\%) & Prams (M)\\ \hline
     EMS-Net \cite{EMS-Net} & 57.29 $\pm$ 1.81 & 22.48 $\pm$ 4.29 & 96.32 $\pm$ 0.15 & \textbf{97.14} $\pm$ 0.13 & 4.94\\
     FAMED \cite{FAMED} & 65.22 $\pm$ 5.89 & 23.67 $\pm$ 4.82 & 94.77 $\pm$ 1.84 & 94.06 $\pm$ 3.32 & 4.90\\
     Labram  \cite{jiang2024large}  & 72.74 $\pm$ 4.07 &41.01 $\pm$ 3.30 & 96.41 $\pm$ 0.47& 96.22 $\pm$ 0.82 & 4.70 \\
      EEGPT \cite{wang2024eegpt}  & 63.16 $\pm$ 3.53 & 25.94 $\pm$ 7.02 & 95.67 $\pm$ 0.59 & 95.62 $\pm$ 0.87 & 4.70 \\
    KAN2EEG \cite{herbozo2025kan}  & 58.98 $\pm$ 2.57 & 17.22 $\pm$ 2.24 &95.04 $\pm$ 1.25& 94.90 $\pm$ 2.26 & 9.40 \\
     SimBA \cite{patro2024simba}  & 69.61 $\pm$ 2.80 & 38.48 $\pm$ 3.41 &96.40 $\pm$ 0.46& 96.35 $\pm$ 0.77 & 25.70 \\
     \hline
     CadeNet (ours) & 81.39 $\pm$ 5.37 & \textbf{51.05} $\pm$ 4.91 & \textbf{96.83} $\pm$ 0.56& 96.50 $\pm$ 0.90 & 4.20 \\
    LV-CadeNet (ours)  &  \textbf{86.32} $\pm$ 2.34 & 46.91 $\pm$ 6.32 & 95.96 $\pm$ 0.97 & 95.05 $\pm$ 1.51 & 4.20 \\ \hline
    \end{tabular} 
 }
\end{table}

Table \ref{tb:results-eeg} presents the performance comparison between our method and baseline approaches for EEG spike classification. It is clear that our lightweight LV-CadeNet with 5.87M parameters already outperforms all baseline methods except Labram-large, a significantly larger pretrained model with 369M parameters. Through hyperparameter optimization following the same tuning strategy applied to CadeNet detailed in the last paragraph of this section, our enhanced LV-CadeNet-large achieves state-of-the-art results, outperforming even Labram-large with improvements of 1.16\% in balanced accuracy, 0.78\% in Cohen's Kappa, and 0.32\% in weighted F1 score. The performance advantage becomes more pronounced when compared with other baseline models. For MEG spike detection, Table \ref{tb:results-meg} shows that our CadeNet and LV-CadeNet approaches exhibit superior performance compared with baseline models on all metrics except accuracy. It should be noted that the accuracy metric cannot reliably reflect model discrimination capability in the MEG spike detection task due to extreme class imbalance in the test set where negative samples significantly outnumber positives. We include accuracy results only for direct comparison with the original baseline metrics. 

The following experimental analysis systematically evaluates both the core methodological innovations and the individual contributions of each component through comprehensive ablation studies in CadeNet.

\subsubsection{Benefits of using the long-view input feature we propose}
In this study a systematic evaluation of our long-view input feature construction methodology was carried out using three complementary experimental analyzes: (1) a cross-architectural robustness assessment; (2) a comparative performance analysis against the conventional extended-window raw signal input method; and (3) a qualitative analysis of the contribution of this feature construction approach.

\begin{table}[t]
\caption{Comparative performance of models with and without long-view input features for the EEG spike classification task.}
\label{tb:lv-eeg}
\centering
\scalebox{0.73}{
    \begin{tabular}{c|c|cccccc}
     \hline
     \multicolumn{2}{c|}{\diagbox[]{Model}{Metric}} & BAcc (\%) & CKap (\%) & WF1 (\%) & Prams (M) & Div\\ \hline
     \multirow{2}{*}{SpikeNet  \cite{ref_spikenet}}  & Base  &   46.34 $\pm$ 2.13  & 47.12 $\pm$ 3.36 & 74.48 $\pm$ 1.40 & \multirow{2}{*}{6.51} & \multirow{2}{*}{$\uparrow$} \\
     ~ & LV & \textbf{58.00} $\pm$ 1.93 &  \textbf{57.98} $\pm$ 1.93 & \textbf{78.61 } $\pm$ 0.98 & ~ \\ \hline
     
     \multirow{2}{*}{Labram  \cite{jiang2024large}}  & Base  &  55.47 $\pm$ 3.35  & 56.28 $\pm$ 4.06 & 78.13 $\pm$ 1.99 & \multirow{2}{*}{5.80} & \multirow{2}{*}{$\uparrow$} \\
     ~ & LV & \textbf{64.87} $\pm$ 2.00  &  \textbf{66.07} $\pm$ 1.43 & \textbf{82.31} $\pm$ 0.71 & ~ \\ \hline

     \multirow{2}{*}{EEGPT \cite{wang2024eegpt}}  & Base  &  49.64  $\pm$ 2.24  & 54.07 $\pm$ 3.19 & 77.54 $\pm$ 1.41 & \multirow{2}{*}{4.70} & \multirow{2}{*}{$\uparrow$} \\
     ~ & LV & \textbf{57.34} $\pm$  2.71  &  \textbf{57.10} $\pm$ 2.79 & \textbf{78.38 } $\pm$ 1.40 & ~ \\ \hline

    \multirow{2}{*}{SimBA \cite{patro2024simba}}  & Base  &  46.96 $\pm$ 1.92 & 49.99 $\pm$ 2.33 & 75.48 $\pm$ 1.02 & \multirow{2}{*}{25.80} & \multirow{2}{*}{$\uparrow$} \\
     ~ & LV & \textbf{57.75 } $\pm$ 1.85  &  \textbf{56.08} $\pm$ 1.85 & \textbf{77.74} $\pm$ 1.21 & ~ \\ \hline
    
    \end{tabular} 
 }
\end{table}

\begin{table}[t]
\caption{Comparative performance of models with and without long-view input features for the MEG spike detection task.}
\label{tb:lv-meg}
\centering
\scalebox{0.73}{
    \begin{tabular}{c|c|cccccc}
     \hline
     \multicolumn{2}{c|}{\diagbox[]{Model}{Metric}} & BAcc (\%) & CKap (\%) & WF1 (\%) & Prams (M) & Div\\ \hline
     \multirow{2}{*}{EMS-Net \cite{EMS-Net}}  & Base  & 57.29 $\pm$ 1.81 & 22.48 $\pm$ 4.29 & 96.32 $\pm$ 0.15 & \multirow{2}{*}{4.94} & \multirow{2}{*}{$\uparrow$} \\
     ~ & LV & \textbf{61.10} $\pm$ 2.20 &  \textbf{29.04} $\pm$ 5.33 & \textbf{96.38} $\pm$ 0.37 & ~ \\ \hline
     
     \multirow{2}{*}{Labram  \cite{jiang2024large}}  & Base  & 72.74 $\pm$ 4.07 &\textbf{41.01} $\pm$ 3.30 & \textbf{96.41} $\pm$ 0.47 & \multirow{2}{*}{4.70} & \multirow{2}{*}{$\uparrow$} \\
     ~ & LV & \textbf{74.43} $\pm$ 3.82  &  40.81  $\pm$ 3.10 & 96.21 $\pm$ 0.61 & ~ \\ \hline

     \multirow{2}{*}{EEGPT \cite{wang2024eegpt}}  & Base & 63.16 $\pm$ 3.53 & 25.94 $\pm$ 7.02 & 95.67 $\pm$ 0.59  & \multirow{2}{*}{4.70} & \multirow{2}{*}{$\uparrow$} \\
     ~ & LV & \textbf{65.00} $\pm$ 3.56  &  \textbf{29.61} $\pm$ 6.38 & \textbf{95.89} $\pm$ 0.67 & ~ \\ \hline

    \multirow{2}{*}{SimBA \cite{patro2024simba}}  & Base  & 69.61 $\pm$ 2.80 & 38.48 $\pm$ 3.41 &96.40 $\pm$ 0.46 & \multirow{2}{*}{25.70} & \multirow{2}{*}{$\uparrow$} \\
     ~ & LV & \textbf{70.98} $\pm$ 3.64 &  \textbf{40.92} $\pm$ 3.90 & \textbf{96.54} $\pm$ 0.45 & ~ \\ \hline

    \end{tabular} 
 }
\end{table}

\begin{figure*}[htbp]
    \centering
    \includegraphics[width=\linewidth]{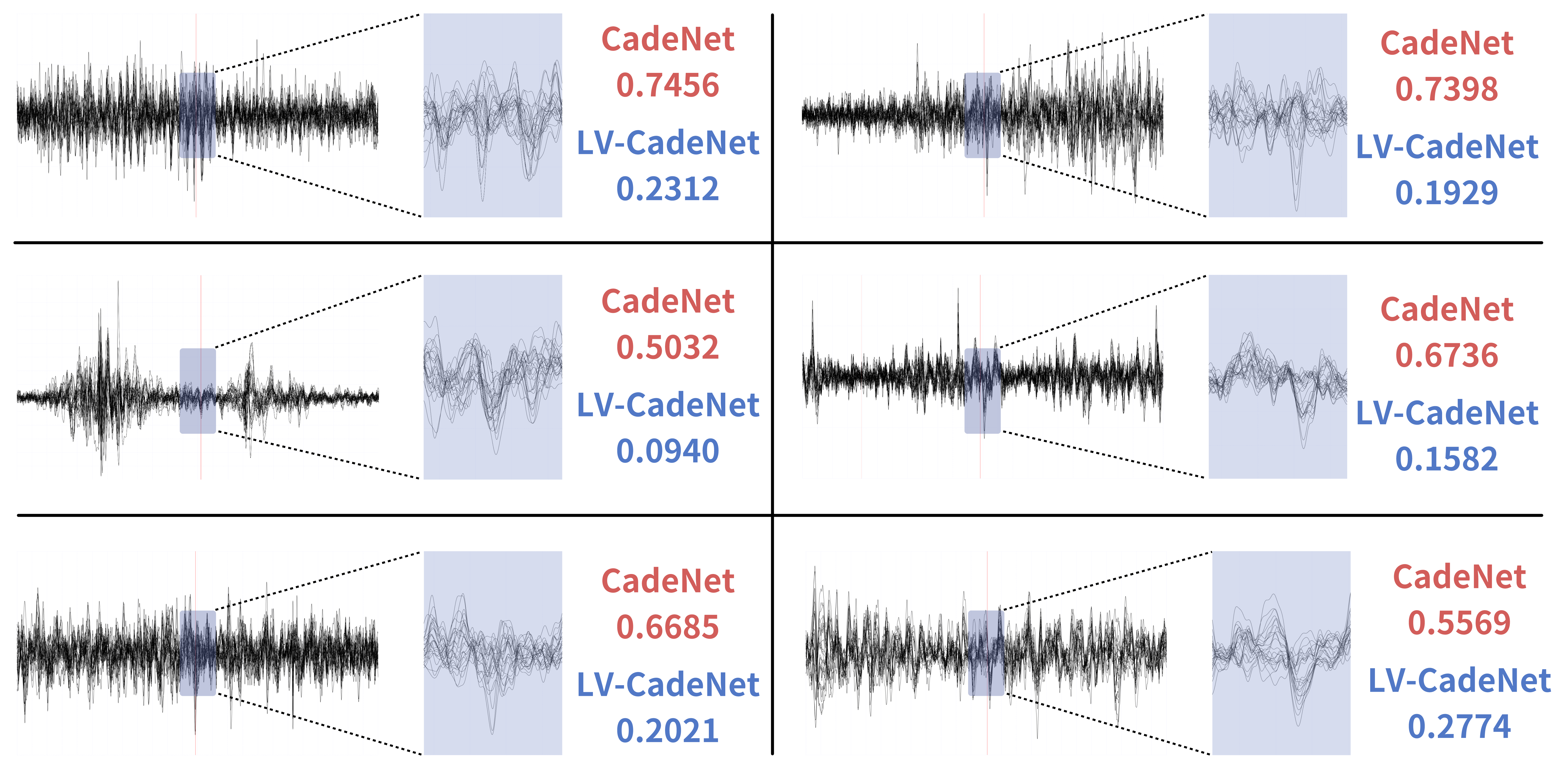}
    \caption{Six exemplary cases where LV-CadeNet correctly identifies non-spike activity while CadeNet fails. Each case shows the 6s MEG signal, a 384ms zoomed-in subsegment, and the models' confidence scores for spike presence in the subsegment.}
    \label{fig:example-longview}
    \vspace{-1.5em}
\end{figure*}

\noindent \textbf{(1) Evaluating the robustness of long-view input features across different network architectures}. 

As evidenced in Tables \ref{tb:results-eeg} and \ref{tb:results-meg}, LV-CadeNet demonstrates consistent performance superiority over CadeNet for both EEG spike classification and MEG spike detection tasks, validating the effectiveness of our long-view feature representation. It is worth noting that our proposed long-view feature construction operates as a versatile plug-and-play module compatible with various network architectures. Comprehensive evaluations in Table \ref{tb:lv-eeg} reveal that integration of this module yields performance gains across all metrics for multiple EEG spike classification models SpikeNet, Labram, EEGPT, and SimBA. Particularly noteworthy is the significant improvement exceeding 8\% in balanced accuracy for all evaluated models. Similarly, Table \ref{tb:lv-meg} demonstrates performance enhancements when applying the long-view features to different MEG spike detection models. For both tasks, we conducted experiments using both domain-specific and domain-shared models to ensure robust validation of our approach across different architectural paradigms. These extensive experimental results collectively establish the robust generalizability and task-agnostic effectiveness of our long-view feature representation across diverse neurophysiological signal processing tasks and model architectures.

\noindent \textbf{(2) Qualitative comparison of our framework with and without long-view input features}. 

To intuitively illustrate the operational mechanism and efficacy of our proposed long-view input feature, Fig. \ref{fig:example-longview} presents six representative cases where LV-CadeNet successfully discriminates non-epileptic spike events that are consistently misclassified by CadeNet in the MEG spike detection task. These examples reveal a fundamental limitation of the conventional approach: its inability to reliably distinguish non-epileptic spikes in short-duration MEG segments due to insufficient contextual information. In these challenging cases, while the waveforms exhibit spike-like morphology, the model fails to accurately assess their diagnostic significance within temporally constrained inputs, leading to false positive classifications of non-epileptic transients as true spikes. In contrast, LV-CadeNet overcomes this limitation through its novel design that synergistically integrates localized signal characteristics with comprehensive morphological context. By normalizing features within a short duration across extended time windows, our method effectively captures the diagnostic significance of waveforms within their broader physiological context. This dual-scale analysis enables more accurate differentiation between epileptiform and non-epileptic events.

\begin{figure}[htbp]
    \centering
    \includegraphics[width=250pt, height=230pt]{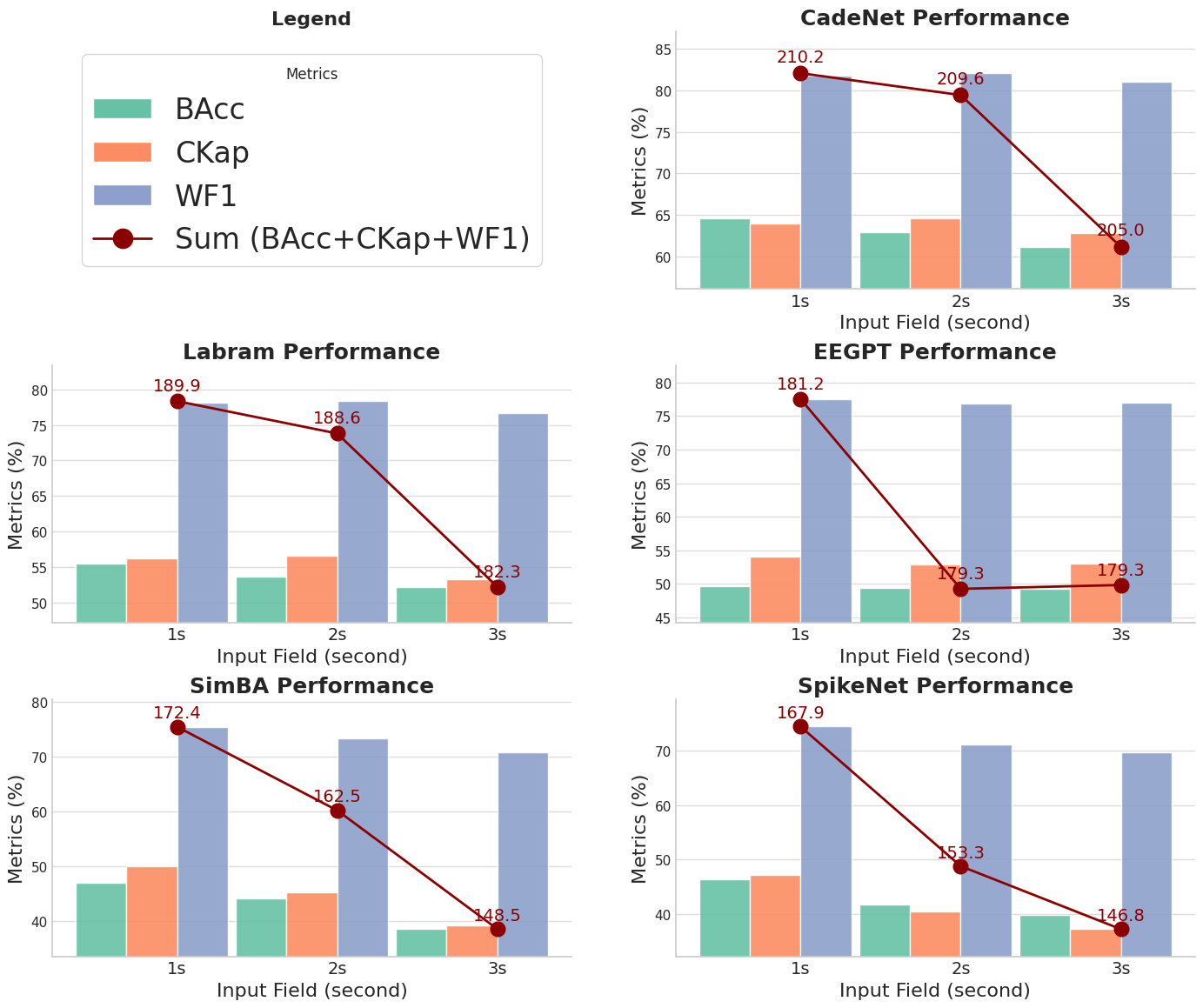}
    \caption{Impact of temporal window extension on model performance using raw signal inputs.}
    \label{fig:input_field}
    \vspace{-1.5em}
\end{figure}

\noindent \textbf{(3) Performance comparison of using our long-view feature versus raw signal input with extended time window}. 

We performed a comprehensive comparative analysis between our long-view feature construction method and the conventional approach of processing raw extended-duration signals for EEG spike classification. As shown in Fig. \ref{fig:input_field}, extending the temporal duration of input signals (input fields) consistently degrades classification performance in all evaluated models, including our CadeNet and the four baseline architectures Labram, EEGPT, SimBA, and SpikeNet. This consistent observation leads us to hypothesize that naive temporal expansion reduces the relative proportion of discriminative feature regions, where any potential benefits from enhanced feature salience are negated by the introduction of excessive noise. These compelling findings unequivocally demonstrate that our long-view feature construction method establishes a new paradigm that is essential to improve spike analysis performance.

\subsubsection{Benefits of the CadeNet architecture we propose} The experimental results presented in \ref{tb:results-eeg} and \ref{tb:lv-meg} demonstrate that our CadeNet consistently outperforms all baseline models in both EEG spike classification and MEG spike detection tasks. This superior performance can be attributed to three fundamental architectural innovations. First, the hybrid convolution-attention module effectively captures the spatiotemporal dynamics of neural signals by integrating convolutional operations for temporal feature extraction with attention mechanisms for spatial feature learning. Second, the multi-scale feature compression framework not only improves computational efficiency but also expands the receptive field of both convolution and attention operations, thereby enabling more effective extraction of higher-level abstract features. Finally, the decoupled encoder-decoder architecture provides a significant advantage over direct classification approaches by employing a dedicated decoder to generate optimized task-specific representations, which substantially enhances the model's discriminative capability. Together, these innovations establish CadeNet as an advanced solution for EEG/MEG spike analysis. We also conduct ablation studies in this subsection to analyze CadeNet's hyperparameter sensitivity and establish the configuration that yields optimal performance.

\noindent \textbf{(1) Individual contributions of the three key architectural innovations.} 

To rigorously evaluate the individual contributions of key factors listed above, we conducted systematic ablation studies on the public TEUV dataset for EEG spike classification. The comprehensive results are presented in Table \ref{tb:cadenet_components}, organized into three critical comparisons. Firstly, replacing fully-connected layers with convolutional operations for temporal feature extraction yields a significant 12.17\% improvement in aggregate performance across all evaluation metrics. This result implicitly validates the superior capability of convolutional architectures in capturing discriminative temporal patterns from neural signals. Secondly, our feature extraction with progressive spatiotemporal downsampling increases the abstraction of the spatial-temporal features (IncreAbs T/S) and demonstrates clear advantages over approaches maintaining equal feature dimensions (EqAbs). As shown in the second block of Table \ref{tb:cadenet_components}, this design achieves 14.93\% gains in aggregated performance, confirming the importance of multi-scale abstraction for robust feature learning.
Finally, the decoder component proves essential for performance, with the third block of Table \ref{tb:cadenet_components} showing 4.54\% aggregated performance gains when comparing CadeNet with versus without this module. These results collectively demonstrate that factor listed in the previous paragraph makes distinct and measurable contributions to the model's classification capability.

\noindent \textbf{(2) Ablation studies on the hyperparameters of CadeNet.} 

We conducted an ablation study to examine the critical hyperparameters of CadeNet, with particular focus on the convolutional operations and decoder architecture. As suggested in the preceding analysis, these components most significantly impact model performance. Specifically, we investigated two key aspects: (1) the receptive field size in convolutional feedforward blocks, and (2) variations in decoder depth. As evidenced by the experimental results shown in the first and third blocks of Table \ref{tb:cadenet_components}, expanding the model's architectural complexity, whether through enlarged convolutional receptive fields or increased decoder depth, fails to yield performance gains. These findings indicate probable overfitting phenomena, most likely stemming from the constrained sample size available in our dataset.

\begin{table}[t!]
\caption{Benefits of the CadeNet architecture we propse.}
\begin{center}
\scalebox{0.7}{

\begin{tabular}{c|ccccc}
\hline
\diagbox[]{Model}{Metric} & BAcc (\%) & CKap (\%) & WF1 (\%) & Sum & Params(M)  \\ \hline

\multicolumn{5}{c}  {\textbf{\textit{Using convolutional operation for temporal feature extraction }}}\\ \hline
    FC  & 56.43 $\pm$  2.10 & 61.20 $\pm$ 2.23 & 80.45  $\pm$  1.16 &  198.08 & 2.47 \\
    Conv-ker3 (CadeNet) & \textbf{64.55} $\pm$ 2.41 & \textbf{63.96} $\pm$ 2.44 & \textbf{81.74} $\pm$ 1.32 & \textbf{210.25} & 5.87 \\ 
    Conv-ker5 & 61.51 $\pm$ 2.28 & 62.96  $\pm$ 2.73 & 81.23  $\pm$ 1.37 & 205.7 & 9.27 \\ 
    Conv-ker7 & 59.11  $\pm$ 3.30 & 60.59 $\pm$ 3.80  &  80.01 $\pm$ 2.03 & 199.71 & 12.67 \\ \hline
    
\multicolumn{5}{c}  {\textbf{\textit{Feature extraction with increasing abstraction}}}  \\  \hline
EqAbs T/S   & 57.10  $\pm$  3.50   & 58.80  $\pm$ 2.69   & 79.42  $\pm$ 1.33   & 195.32 & 5.31 \\
IncreAbs S  & 60.91 $\pm$ 3.40 &  61.07  $\pm$ 3.99   & 80.22 $\pm$ 2.22   & 202.20 & 5.52 \\
IncreAbs T & 61.83  $\pm$ 1.98 &   62.79 $\pm$  1.75  & 81.11 $\pm$ 0.94   & 205.73 & 5.23 \\
IncreAbs T/S (CadeNet)  & \textbf{64.55} $\pm$ 2.41   & \textbf{63.96} $\pm$ 2.44   & \textbf{81.74} $\pm$ 1.32   & \textbf{210.25} & 5.87 \\  \hline
\multicolumn{5}{c}  {\textbf{\textit{Integration of the decoder}}}  \\  \hline
 w/o decoder-S  & 60.12   $\pm$ 2.22 & 63.87  $\pm$ 2.17 & 81.72  $\pm$  1.11 & 205.71 & 3.62 \\
 w/o decoder-L  & 60.92   $\pm$ 1.89 & 63.76   $\pm$ 1.88 & 81.72   $\pm$  0.98 & 206.4 & 5.87 \\
 w/ decoder-2B (CadeNet) & \textbf{64.55} $\pm$ 2.41 & \textbf{63.96} $\pm$ 2.44 & \textbf{81.74} $\pm$ 1.32 & \textbf{210.25} & 5.87 \\ 
 w/ decoder-4B  & 62.48  $\pm$ 2.45 & 63.25 $\pm$ 2.30 & 81.30 $\pm$ 1.22 & 206.96 & 8.12 \\ 
 w/ decoder-6B  & 61.49  $\pm$ 3.59 & 62.80 $\pm$ 2.45 & 81.13  $\pm$ 1.26 & 205.42 & 10.37 \\ 
 \hline

\end{tabular} 
}
\begin{flushleft}
    \justifying
      \footnotesize{The notation `FC' and `Conv' denote temporal feature extraction using fully connected and convolutional layers respectively; `EqAbs' represents spatiotemporal feature extraction with equal abstraction across dimensions while `IncreAbs' indicates spatiotemporal feature extraction with progressively increasing abstraction. `S' and `T' refers to the temporal and spatial dimensions, respectively. The notation `w/o decoder-S' refers to removing all decoder blocks and performing classification directly using the encoder's output features. In contrast, `w/o decoder-L' indicates relocating the two decoder blocks to the encoder before classification, ensuring comparable model complexity between configurations with and without a decoder for fair comparison. The notation `w/ decoder' is followed by the number of blocks retained in the decoder.
    }      
    \end{flushleft}
\label{tb:cadenet_components}
\end{center}
\vspace{-1.5em}
\end{table}

\section{Conclusion and Future Work}
This study presents LV-CadeNet, an innovative deep learning framework designed to advance EEG/MEG spike analysis through two fundamental breakthroughs. First, we develop a novel long-view input feature representation that comprehensively captures both transient spike characteristics and their extended neurophysiological context, overcoming the limited temporal perspective of conventional approaches. Second, we architect a sophisticated convolution-attention fusion encoder-decoder network that synergistically processes spatiotemporal patterns through: (i) multi-scale convolutional-attention layers extracting hierarchical features with progressive abstraction, and (ii) an intelligent decoder that optimally distills discriminative features for precise spike characterization. Comprehensive evaluations demonstrate LV-CadeNet's exceptional performance, establishing new state-of-the-art benchmarks with a weighted F1 score of 83.35\% on the TUEV dataset and remarkable 96.83\% weighted F1 score on clinical MEG recordings. These outstanding results position LV-CadeNet as a promising clinically solution for reliable epileptiform discharge analysis.

While LV-CadeNet has demonstrated excellent performance in spike detection and classification tasks as shown in Tables \ref{tb:results-eeg} and \ref{tb:results-meg}. We recognize that evaluating cross-subject generalizability remains an essential next step for clinical translation. Our future work will systematically investigate this through three complementary approaches: first, implementing rigorous leave-one-subject-out validation protocols across multiple clinical centers to evaluate population-level robustness; second, conducting detailed subgroup analyses to identify potential performance variations across demographic (age, sex) and clinical (epilepsy type, medication status) factors; and third, developing specialized domain adaptation techniques to enhance model generalizability while preserving diagnostic accuracy.

% \section*{References}
% \vspace{-1.5em}
\bibliographystyle{IEEEtran}
\bibliography{Arxiv}
\end{document}